# Generalizable Prediction Model of Molten Salt Mixture Density with Chemistry-Informed Transfer Learning

*Julian Barra[1], Shayan Shahbazi[2], Anthony Birri[3], Rajni Chahal[3], Ibrahim Isah[1], Muhammad Nouman Anwar[1], Tyler Starkus[2], Prasanna Balaprakash[3], Stephen Lam[1]\**

[1]Department of Chemical Engineering, University of Massachusetts Lowell, Lowell, MA-01854, USA

[2]Argonne National Laboratory, Lemont, IL-60439, USA

[3]Oak Ridge National Laboratory, Oak Ridge, TN-37830, USA

*Stephen Lam (Stephen_Lam@uml.edu)

**ABSTRACT.** Optimally designing molten salt applications requires knowledge of their thermophysical properties, but existing databases are incomplete, and experiments are challenging. Ideal mixing and Redlich-Kister models are computationally cheap but lack either accuracy or generality. To address this, a transfer learning approach using deep neural networks (DNNs) is proposed, combining Redlich-Kister models, experimental data, and ab initio properties. The approach predicts molten salt density with high accuracy ($r^2 > 0.99$, MAPE $< 1\%$), outperforming the alternatives.

**MAIN TEXT.** Molten salts are liquid mixtures of cations and anions which have recently commanded widespread interest in their use as energy materials. Owing to their desirable thermophysical properties, such as high heat capacity, and low vapor pressure[1], their potential applications include advanced nuclear reactors, thermal energy storage, pyrochemical reprocessing of used nuclear fuel, carbon capture, and molten salt batteries.[2–5] Their wide applicability across a range of clean energy areas has raised the need for data on their material properties for the purposes of design and optimization. However, obtaining data solely through experiments involves challenging measurements at very high temperatures for salts that are often toxic and very corrosive. Also, given that the experiments involve high costs in both equipment and labor, the massive compositional space of molten salts make their exploration through experimentation alone impractical. These issues have driven the use of alternative methods for modeling and evaluating molten salt properties, including molecular dynamics (MD) for atomistic simulations to evaluate material properties.[6] Here, while the MD approach has proven successful,[7–16] it involves a high computational cost and thus, is unfeasible for the prediction of more than a small number of systems and properties. As such, the search for computationally cheaper methods to efficiently navigate the temperature and compositional space of molten salts remains critical for the timely development of next-generation molten salt technologies.

The semi-empirical Redlich-Kister (RK) method was recently demonstrated for predicting molten salt mixture density ($\rho_{mix}$) and viscosity ($\mu_{mix}$) for up to 3 pseudo-components using equations fitted to data available in the Molten Salt Thermal Properties Database-Thermophysical (MSTDB-TP).[17,18] The MSTDB-TP contains property correlations fitted from over 140 published studies for thermophysical properties of interest in molten salts, which include macroscopic density, viscosity, thermal conductivity, and heat capacity. The database contains properties on over 448 different molten salt systems ranging from pure salts (e.g., LiF), to mixtures of up to four pseudo-components (e.g., LiF-NaF-KF-UF$_4$) with material densities being the most well-characterized property. This is due to its fundamental importance in relation to thermodynamic properties and relative ease and accuracy of measurement. The empirical equation for macroscopic density as a function of temperature for a given system in MSTDB-TP is shown below:

$$\rho(T) = A - B \cdot T \tag{1}$$

Where A and B are fitted coefficients for each composition.[15] This equation can be used to generate density data across temperature, which RK expansions can be used to interpolate between both composition and temperature, calculating mixture density as the sum of the density calculated under ideal mixing assumptions and an excess density representing the deviation from ideality due to pseudo-component interaction effects (Equation 2). Ideal density is calculated according to Equation 3 and the excess density is calculated according to Equations 4 and 5:

$$\rho_{mix} = \rho_{id} + \rho_{ex} \tag{2}$$

$$\rho_{id} = \left(\sum_{i=1}^{S} x_i M_i\right) \sum_i^S \frac{x_i M_i}{\rho_i} \tag{3}$$

$$\rho_{ex} = \sum_{a=1}^{S-1} \sum_{b=2}^{S} x_a x_b \sum_{j=1}^{n} L_j^{ab} (x_a - x_b)^{j-1} \tag{4}$$

$$L_j^{ab} = A_j^{ab} + B_j^{ab} T \tag{5}$$

Where $S$ is the total number of components in the system, $x_i$ and $M_i$ are the molar concentration and molar mass of component $i$ with experimentally measured density $\rho_i$, $x_a$ and $x_b$ are the molar concentrations of components $a$ and $b$ which are considered a pseudo-binary subsystem of the system of size S, and $L_j^{ab}$ are linear temperature-dependent interaction parameters associated with pseudo-binary subsystem components $a$ and $b$. The coefficients in $L_j^{ab}$ are fitted to available experimental data (typically pseudo-binary, ternary or both). As such, RK expansions improve the accuracy of ideal mixing model (Equation 3) by accounting for non-ideal contributions with system-specific parameters (Equation 4 and 5). Herein, the RK formalism in Equations (2)-(5) can be used to 1) interpolate between experimental data points collected for lower order mixtures (typically pseudo-binary or pseudo-ternary) at different temperatures and compositions, or 2) to extrapolate to higher-order mixture using the coefficients corresponding to the pairings of all compounds present in that mixture (typically fitted from data of pseudo-binary systems). Thus far, RK expansions have been used to perform both kinds of property predictions, and in both cases the approach shows higher accuracy than assuming ideal mixing, reflecting the demonstrated non-ideality in molten salt behavior resulting from interactions between pseudo-components[19]. However, a key limitation of such semi-empirical approaches is in cases where models have not been developed for lower order (binary or ternary) to inform on higher order mixtures, or when data is sparse which can lead to significant interpolation error, particularly towards increasingly non-ideal systems (e.g., $LiF-NaF-ZrF_4$).[20,21]

The present work uses density to demonstrate proof of concept for generalizable deep learning-based property prediction in liquid mixtures. The method proposed here aims to further improve on the accuracy and transferability of RK expansions at a similar computational cost by using MSTDB-TP data to train deep neural networks for interpolating properties across a wide range of compositions and temperatures. While machine learning (ML) methods have been previously used for predicting the properties of materials,[22–26] including ionic liquids,[27–30] compounds, and simple molecules,[31,32] the prediction of mixture properties is scarce.[33] The limited applications of ML for liquid mixtures is likely owed to challenges in representing multiple components in dynamically disordered systems that lack defined structure in the solution phase.

In molten salts, ML has largely been used to learn the potential energy surface with interatomic potentials used in MD simulations, thereby learning the atomic energies as a function of the evolving local chemical environment for a given system.[8–10,34] Recently, ML prediction of molten salt properties based on simple atomistic descriptors was demonstrated for a simple property (melting point) and a specific class of salts (mixed alkali halide reciprocal salts)[35]. However, efficiently predicting the properties of all molten salt mixtures directly in a generalized manner based on their composition and thermodynamic conditions is challenged by lack of large datasets that are typically required for training deep learning models. To deal with these challenges, this work proposes a methodology for the prediction of molten salt properties that addresses these concerns about 1) model transferability (due to a relatively small training set limited by experimental data), through transfer learning (TL)[36] of insights from state-of-the-art semi-empirical relations (i.e., RK expansion); and 2) generalizability, by encoding molten salt mixtures using the fundamental chemical properties of its components as descriptors. The approach for generalizability is based on theory of molten salt behavior which proposes a relationship between salt mixture properties and atomistic features of the components within the salt mixture such as charge (or electronegativity) and ionic radius (or polarizability)[35,37,38]. A key objective of this workflow is to create models to predict salt mixture properties with a higher accuracy than RK expansions, while retaining generalizability of ideal or semi-empirical non-ideal mixing models, at a similarly low computational cost.

First, a dataset is created using all available compositions from the MSTDB-TP (Thermophysical) v3.0[39] by sampling the density in Equation 1 at 50 K intervals across the applicable temperature range (ATR) for each system. The temperature and molar fractions of the compounds in the mixture are added as descriptors. To make the inputs for chemical species generalizable across the periodic table, chemical information of the salts is encoded using JARVIS Classical Force Field Inspired Descriptors (JARVIS-CFID)[40] database of ab-initio calculations, which contains 1557 chemical-structural-charge descriptors for any compound (e.g., LiF) at 0 K. These descriptors are further down-selected; details of the down-selection process are provided in the supplementary information, along with the full list of the final descriptors. A general DNN model is developed for up to 4 components by concatenating descriptors of each component compound and using zero-padding to represent lower order systems in the input. After down-selection, a total of 134 total descriptors are added from JARVIS-CFID for each component. All chemically equivalent orderings of the input are added to the dataset (e.g., for a pseudo-binary mixture, both $[\mathbf{C_1}, \mathbf{C_2}]$ and $[\mathbf{C_2}, \mathbf{C_1}]$ are included where $\mathbf{C_1}$ and $\mathbf{C_2}$ are the set of descriptors for components 1 and 2, respectively) to ensure permutational invariance. The final 'experimental' dataset contained 52,320 data points. A test set consisting of 20% of the overall dataset is withheld during training and used to check model performance on unseen data.

To improve transferability, reduce overfitting in data-sparse regimes, and regularize the DNN model, a transfer learning (TL)[41] process is implemented as shown in Figure 1, in which an approximate DNN is initially trained with large quantities of Redlich-Kister modeled data sampled across temperature and composition. This approximate DNN model is then refined with the smaller experimental dataset sampled from MSTDB-TP. During TL, the first two hidden layers of the initial DNN are retained and used to create a new network by adding three new layers with an additional output neuron. As shown in Figure 1(c-d), this new DNN is then trained on the MSTDB-TP (comparatively smaller experimental dataset) two times – first with the first two layers frozen to retain the information of the old model, and then a second time with all layers unfrozen using a very small learning rate ($\eta =$1E-6) to fine-tune it. By preconditioning the model with large quantities of RK data, approximate relationships can be learned between the descriptors and mixture density, and the need for benchmarked experimental data from MSTDB-TP can be minimized. The RK dataset is generated using the available parameters for binary RK expansions given in the released version of the MSTDB-TP v.3.0[39]. Density data is created for the pseudo-ternary systems in the MSTDB-TP, at increments of 10% molar fraction for each compound, and including all chemically equivalent orderings of components in each data point. The chemical information of the salts is encoded using JARVIS-CFID descriptors as done for the experimental dataset. The final RK dataset used to train the DNN consists of 134,784 data points.

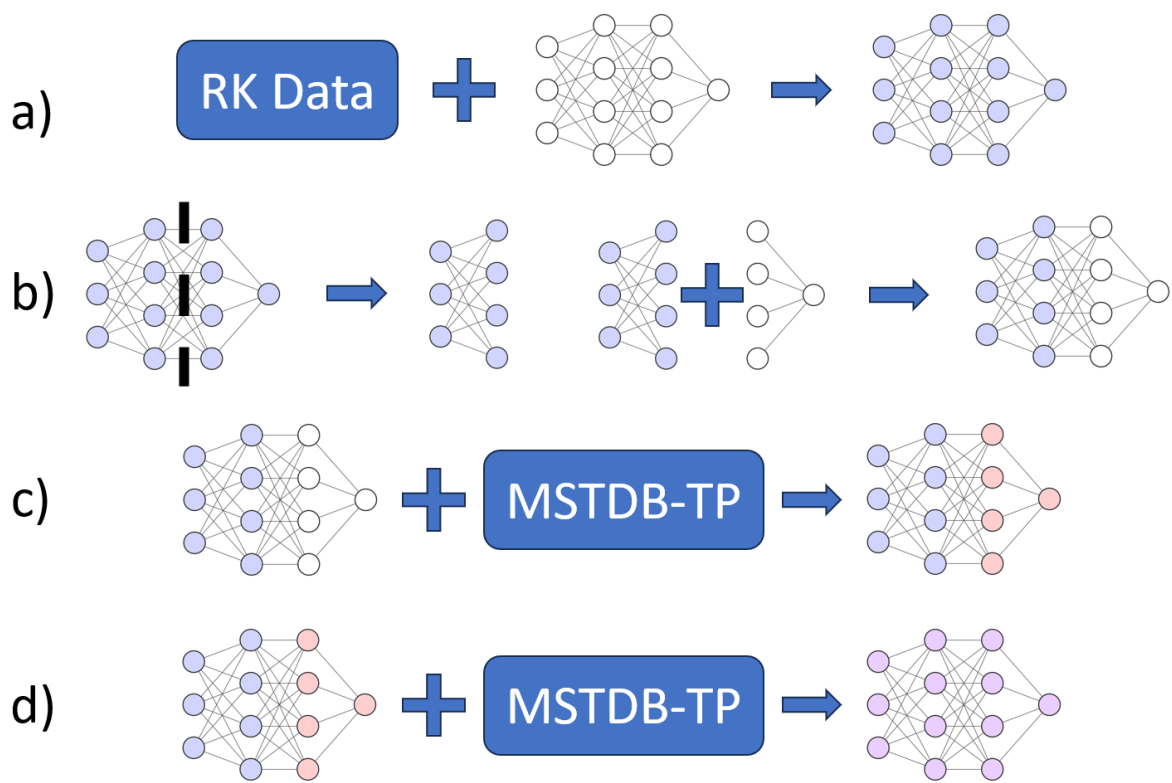

**Figure 1.** Transfer learning (TL) workflow performed whereby knowledge from RK models are transferred to the DNN and updated with a smaller experimental dataset. The steps are as follows: a) The DNN is trained on density data generated across temperature using binary RK expansions.

b) The latter half and the output neuron of the trained model are discarded, and the first two layers are frozen and used to create a new model by adding untrained layers and an output neuron. c) The new model is trained on the dataset generated with the MSTDB-TP. d) The model is trained again on the dataset generated with the MSTDB-TP, this time with all layers unfrozen.

As such, the DNN trained on the RK dataset is composed of four hidden layers with 128 neurons each, whereas the version trained on the MSTDB-TP data is composed of five hidden layers (two taken from initial model plus three new layers). The networks are implemented with L2-norm regularization and the number of nodes is tuned by minimizing the loss on a validation set consisting of a randomly selected 20% of the training set at each epoch.

In Figure 2, the performance of RK is compared to DNN for predicting the density of pseudo-binary (Figure 2a), and pseudo-ternary (Figure 2b) molten salts in the MSTDB-TP with mean absolute error (MAE), mean absolute percentage error (MAPE), and $r^2$ compared in Table 1. Here, the best available RK model is used for each system. Namely, the pseudo-ternary systems are fitted with system-specific pseudo-ternary interaction parameters for comparison.

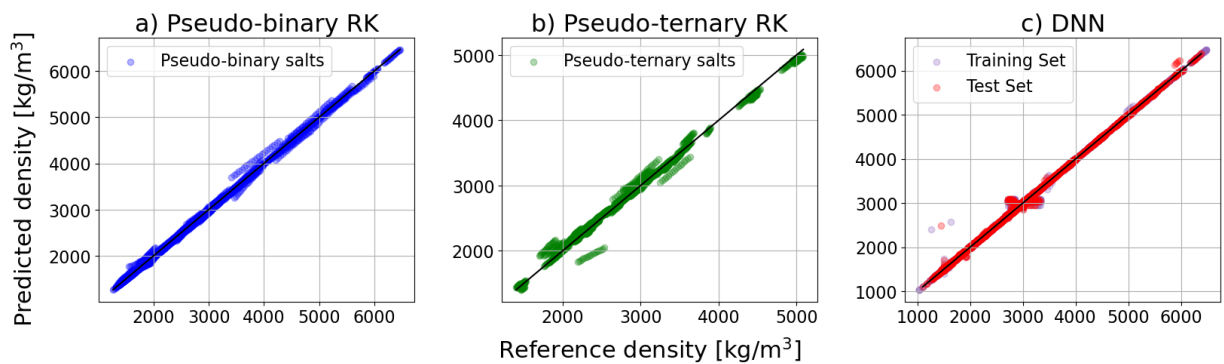

**Figure 2.** Parity plots (predicted vs. reference values) for density [kg/m³] as predicted by a) RK expansions for pseudo-binary mixtures using binary interaction coefficients, b) RK expansions for pseudo-ternary mixtures using ternary interaction coefficients, and c) a deep neural network across the entire MSTDB-TP dataset containing both pseudo-binary and pseudo-ternary mixtures.

**Table 1.** Mean absolute error (MAE), mean absolute percentage error (MAPE), and coefficient of determination ($r^2$) calculated on the predictions made by a deep neural network (DNN), RK expansions using binary interaction coefficients (Binary), and RK expansions using ternary interaction coefficients (Ternary).

|  | Overall | | |
| --- | --- | --- | --- |
|  | **DNN** | **RK Binary** | **RK Ternary** |
| **MAE** $\left[\frac{kg}{m^3}\right]$ | 5.593 | 20.87 | 43.10 |
| **MAPE** [%] | 0.213 % | 0.768 % | 1.711 % |
| $r^2$ [−] | 0.9993 | 0.9987 | 0.9900 |

The results show the higher overall accuracy of the DNN, which predicts density with a MAE of 5.593 kg/m$^3$ and a MAPE of 0.213%, which is significantly lower than the predictions made by RK for binary systems (MAE=20.87 kg/m$^3$, MAPE=0.768 %) and for ternary systems (MAE=43.10 kg/m$^3$, MAPE=1.711%). In ternary systems, RK exhibits higher error due to increased complexity of pair interactions between the mixture components, which is readily overcome by the increased expressiveness afforded by the DNN model. Furthermore, while ternary RK models require significant amounts of data from the system's lower order mixtures, the DNN can learn from the entire database and interpolate relationships in other systems containing the same or chemically similar component salts. Furthermore, the DNN achieves an error below experimental uncertainty, which typically ranges from 1-2% and can be as high as 6% for the higher-order mixtures in the MSTDB-TP at the lowest temperatures[42]. Newer methods developed for pycnometric density measurement in molten salts achieve an overall error of 0.3%[43], which is higher than the error achievable by DNN, but not by RK for neither binary nor ternary mixtures. The more widely used Archimedean method can achieve uncertainties almost as low, of 0.5%, still under RK and Ideal mixing uncertainties, yet over DNN uncertainty. As such, well-validated DNNs could potentially be used to benchmark experimental measurements and methods.

To further assess model performance (smoothness, physical correctness, predictability), the ability of the DNN to capture the dependence of density on temperature and composition is examined in more detail for a variety of prototypical systems. In Figure 3, DNN-based $\rho(T, x)$ predictions for $LiF-BeF_2-ZrF_4$, $NaF-LiF-BeF_2$, $LiF-BeF_2-ZrF_4$, and $NaF-ZrF_4-UF_4$ are shown and compared to the ideal and RK models recently developed in Ref[20] containing binary interaction parameters (Figure 3a) and ternary interaction parameters (Figure 3b). The standard error metrics (MAE, MAPE, $r^2$) are calculated for all prediction models across 100 temperature samples within the applicable temperature range in the MSTDB-TP and shown in Table 2.

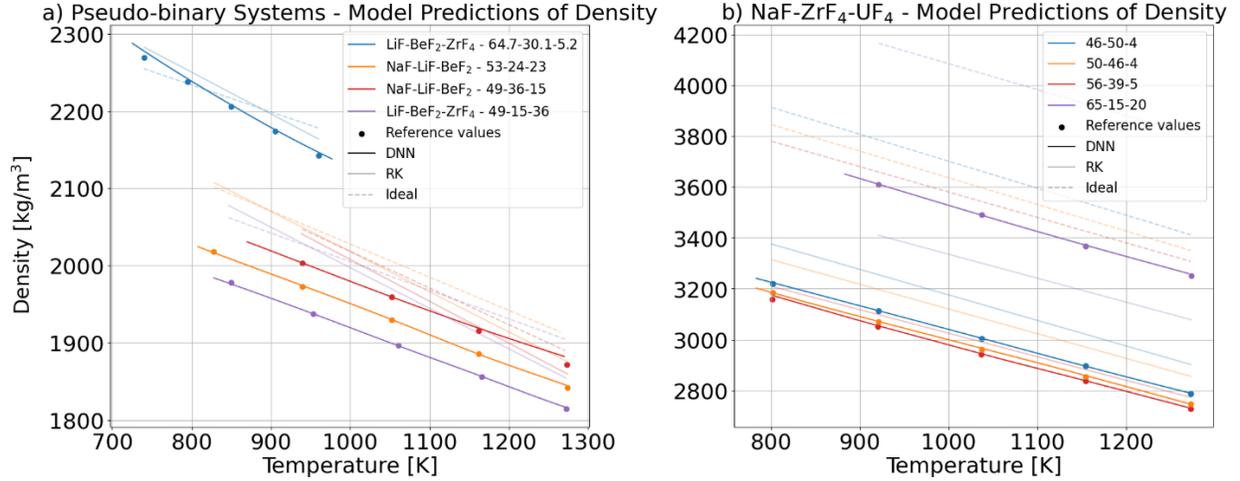

**Figure 3.** Molten salt density versus temperature for a) LiF-BeF$_2$-ZrF$_4$, NaF-LiF-BeF$_2$, and LiF-BeF$_2$-ZrF$_4$ compositions using RK expansion with pseudo-binary interaction parameters, and b) NaF-ZrF$_4$-UF$_4$ using RK expansion with ternary interaction parameters.

**Table 2.** Mean absolute error (MAE) and mean absolute percentage error (MAPE) calculated for the predictions made by the deep neural network (DNN), the binary Redlich-Kister expansion (RK) and ideal mixing (Ideal) on the systems shown in Figure 3.

|  | **Overall** | | |
|---|---|---|---|
|  | **DNN** | **RK** | **Ideal** |
| **MAE** $\left[\frac{kg}{m^3}\right]$ | 3.960 | 80.50 | 334.2 |
| **MAPE** [%] | 0.151% | 2.938% | 9.708 % |

Here, DNN exhibits the lowest error with MAE = 3.960 kg/m$^3$ and MAPE = 0.151% compared to RK models, which have MAE = 80.50 kg/m$^3$ and MAPE = 2.938 %, and ideal mixing which has MAE = 334.2 kg/m$^3$ and MAPE = 9.708 %. As shown in Figure 3, the DNN more accurately captures both the temperature and composition dependence of density in molten salts across a range of salt systems, and models appear to be well regularized. While the ideal model accurately predicts the order of magnitude of LiF-BeF$_2$-ZrF$_4$ density, the temperature-derivative $\frac{\partial \rho}{\partial T}$ (and therefore thermal expansion coefficient $\alpha_V$) is under-predicted, resulting in density overprediction above 820K and underprediction below 820K. Further, the ideal mixing model overpredicts density of 53-24-23 LiF-NaF-BeF$_2$ by 75.51 kg/m$^3$, the density of 49-36-15 LiF-NaF-BeF$_2$ by 30.77 kg/m$^3$, the density of 49-15-36 NaF-KF-BeF$_2$ by 86.57 kg/m$^3$. As such, while ideal models can be used to make predictions even when there is no experimental data, it is clear that the performance of the ideal mixing model is strongly dependent on the system-specific chemical interactions which cannot be known a priori. While the RK expansions generally improve the prediction compared to ideal mixing, the improvements are not systematic. For example, RK-predicted $\left|\frac{\partial \rho}{\partial T}\right|$ can differ significantly from experiments (+45% for NaF-LiF-BeF$_2$ and -10% LiF-BeF$_2$-ZrF$_4$) as shown in Figure 3a. Meanwhile for LiF-NaF-UF$_4$, density is significantly underpredicted ($\Delta\rho\sim$200 kg/m$^3$) for 65-15-20 LiF-NaF-UF$_4$, but overpredicted ($\Delta\rho\sim$180 kg/m$^3$) for 46-50-4 LiF-NaF-UF$_4$ using RK ternary coefficients in Figure 3b. As such, the magnitude and direction of errors are highly dependent on system, composition and temperature, and are difficult to predict.

Meanwhile, the DNN accurately interpolates density across temperature. The accuracy of the model when interpolating across composition is more important to ascertain, given the scarcity of data and impossibility of sampling across molar fraction that has led to the implementation of the TL workflow. To showcase the accuracy of the DNN model across composition when compared to the alternative models and the effectiveness of the TL process, density predictions across varying composition of NaF-LiF-ZrF$_4$ and LiF-BeF$_4$-ThF$_4$ (challenging systems identified in Ref[20]) are shown in Figure 4. Figures 4a and 4b show density predicted across the molar fraction of $ZrF_4$, and Figures 4c and 4d show predictions across the ratio between molar fractions of $BeF_2$ and $ThF_4$ in their corresponding systems. The plots are reproduced along with predictions made across these data-sparse systems by two different DNN models: one trained through the TL workflow previously outlined and the other one trained on the MSTDB-TP generated dataset directly, to show the effectiveness of the TL workflow. The overall metrics for the predictions calculated using the reference values of all available compositions are shown in Table 3.

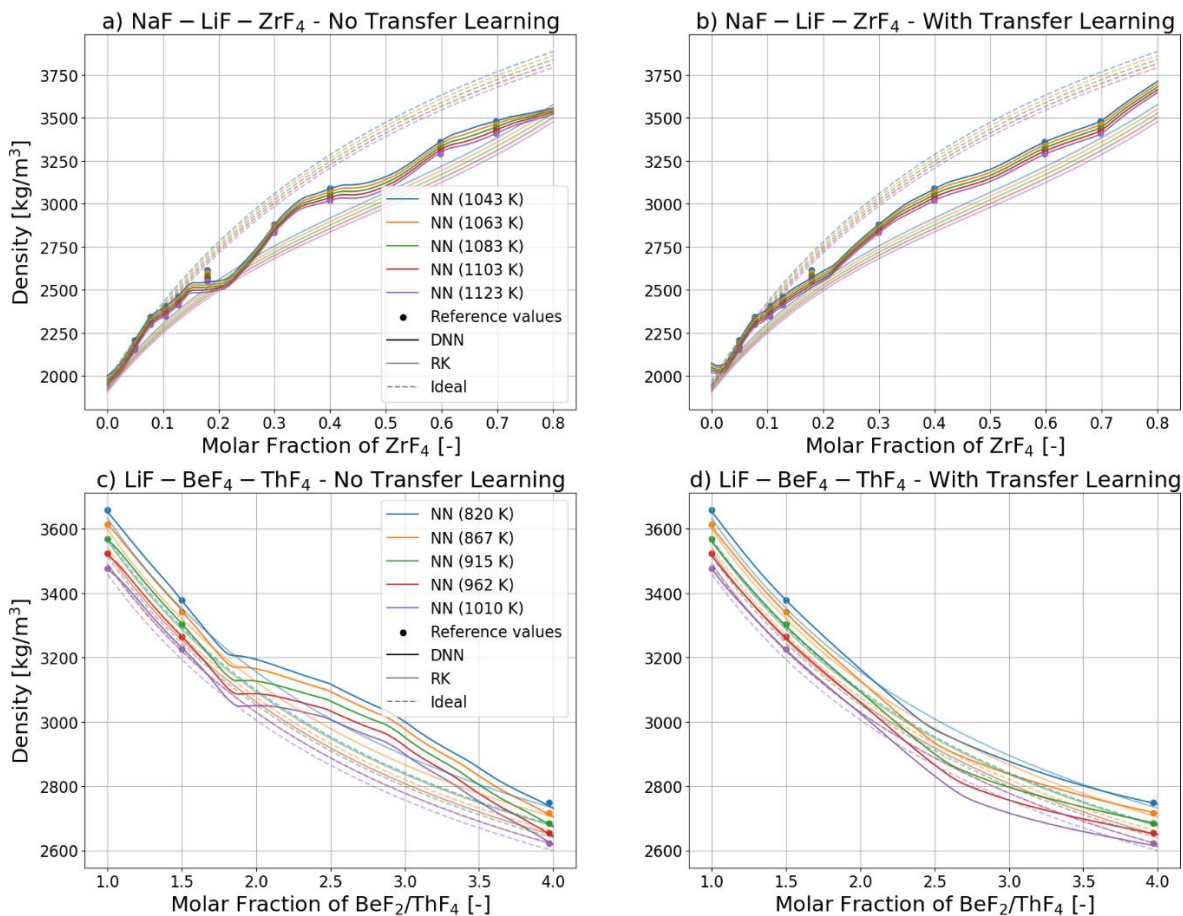

**Figure 4.** Reference values (points) and predictions of molten salt density by DNN (solid lines), RK (semi-transparent lines), and ideal mixing (dashed semi-transparent lines) models of a) NaF-LiF-$ZrF_4$ across molar fraction of $ZrF_4$, with NaF and LiF in a fixed 2:3 molar ratio and DNN

predictions made by a model trained directly on MSTDB-TP data, b) The same system as in 4a, but with DNN predictions made by a model created through transfer learning, c) LiF-BeF$_2$-ThF$_4$ across the molar fraction ratio between BeF$_2$ and ThF$_4$, with LiF fixed at a constant 70% mol, and with DNN predictions made by a model trained directly, and d) the same compositions as in 4c, but with DNN predictions made by a TL-trained model. In all cases, RK expansions include parameters modeling up to pseudo-binary material interactions.

**Table 3.** Mean absolute error (MAE) and mean absolute percentage error (MAPE) calculated for the predictions made by the deep neural network (DNN) with transfer learning, the binary Redlich-Kister expansion (RK) and ideal mixing (Ideal) on the systems shown in Figure 4.

|  | Overall | | |
|---|---|---|---|
|  | **DNN** | **RK** | **Ideal** |
| **MAE** $\left[\frac{kg}{m^3}\right]$ | 8.831 | 87.27 | 107.3 |
| **MAPE** [%] | 0.332 % | 3.142 % | 3.537 % |

As shown in Figure 4, the density increases with a higher proportion of component with relatively higher molar weight (ZrF$_4$ in the NaF-LiF-ZrF$_4$ and ThF$_4$ in LiF-BeF$_2$-ThF$_4$) as expected. The DNN predicts with a MAE of 8.831 kg/m$^3$ and a MAPE of 0.332 %, a significantly lower error than that attained by RK (MAE of 87.27 kg/m$^3$, MAPE of 3.142 %) and ideal mixing (MAE of 107.3 kg/m$^3$, MAPE of 3.537 %). This is particularly salient in the case of NaF-LiF-ZrF$_4$, where the ideal and RK models diverge from the experimental data towards higher mole fractions of ZrF$_4$ as shown in Figure 4. Ideal mixing overpredicts density in NaF-LiF-ZrF$_4$ system and underpredicts it in LiF-BeF$_2$-ThF$_4$, whereas RK expansions do the inverse. Although DNNs can achieve a high overall error without TL (Figure 4a and 4c), models exhibit evidence of overfitting with large fluctuations between the experimental data points. Meanwhile, TL-trained models predict $\rho$ and $\frac{\partial \rho}{\partial x}$ more smoothly, showing the effectiveness of the TL workflow in regularizing the model to make physically realistic predictions, especially for systems with limited data (as was the case with LiF-BeF$_4$-ThF$_4$). With TL, the NN predicts density with a higher overall accuracy, and it better captures both temperature and compositional dependence when compared to the alternative models. This shows that TL provides a useful methodology for creating ML models that can better interpolate databases and explore regions where experimental data is absent. This is critical for many salt applications where compositions and thermodynamic conditions can change during operation (e.g., fission), and where a large design space exists for optimizing salt properties.

In examples presented in Figures 3 and 4, and the overall dataset presented in Figure S1, we have shown significant challenges in modeling chloride and fluoride salts using existing modeling methods. While the density pure symmetric salts (e.g., alkali metal halides) can be modeled via the additive molar volume of their constituents,[44,45] relationships have not been established for non-symmetric salts, and even less so for salt mixtures that exhibit strong non-idealities that affect density in ways that cannot be systematically predicted. While empirical correlations based on ionic radii, charge and quasichemical models have been used for modeling salt density,[44,46,47] such approaches still rely on system-specific fitted parameters, and universal correlations cannot be uncovered. Here, generalized deep learning models are used to learn the underlying relationships between chemical-structural features of salt constituents and its resultant mixture density across 448 pure, pseudo-binary, and pseudo-ternary fluoride and chloride salt systems. Across the dataset of chlorides and fluorides (Section S4, Figure S2 and Figure S3 in Supporting Information), we have demonstrated that this method is robust, and that density, a key thermodynamic property, can be successfully correlated to underlying atomic properties across the periodic table. This presents promising opportunities for developing new theories and significantly improving our fundamental understanding of molten salt properties.

In summary, a chemistry-informed approach for training a transferable and generalizable DNN was developed and shown to accurately predict density in molten salts across a wide composition-temperature space for the first time. The model predicts density of salts in the MSTDB-TP with an overall MAPE of 0.213%, a higher accuracy than that achieved by RK on pseudo-binary systems (0.768%) and RK on pseudo-ternary systems (1.711%). The error achieved by the DNN is within experimental uncertainty, demonstrating accuracy that might be only limited by the inaccuracies in the experimental methods used to acquire training data. Using the DNN on data sparse molten salt systems shows it can smoothly and accurately predict density at a high resolution of temperature and compositional ratios significantly better than state-of-the-art semi-empirical models, which would require large datasets of subcomponent mixtures, and is limited by functional form for capturing complex interactions. While the semi-empirical models are still necessary to generate data for the TL process, the DNN can predict density for systems for which subcomponent data is not available. The DNN also predicts density with a higher accuracy than the ideal models, which do not require as much subcomponent information but perform significantly worse due to an inability to account for significant contributions from excess density. These qualities make the DNN model suitable for interpolating the available data for density in molten salts, indicating that this approach can be extended for the prediction of other critical molten salt properties for which information is available, such as viscosity and heat capacity. Furthermore, the developed method can be applied to studying a wide range of materials mixtures including mixed oxides for thermal energy storage and energy conversion systems, nitrate molten salts used in hydrocarbon mixtures, thermal energy storage, and many more clean energy applications, although further work must be done to assess the transferability of the model towards these salts and to other materials.[48–52]

**Author information**

**Corresponding author**

**Stephen Lam -** University of Massachusetts Lowell, Department of Chemical Engineering, 1 University Ave, SOU-203F, Lowell, Massachusetts, United States of America; orcid.org/0000-0002-7683-1201; Email: Stephen_Lam@uml.edu.

**Authors**

**Julian Barra -** University of Massachusetts Lowell, Department of Chemical Engineering, 1 University Ave, SOU-202E, Lowell, Massachusetts, United States of America; orcid.org/0009-0008-2834-2973; Email: Julian_BarraOtondo@student.uml.edu.

**Shayan Shahbazi –** Argonne National Laboratory, Lemont, Illinois, United States of America; Email: sshahbazi@anl.gov.

**Anthony Birri –** Oak Ridge National Laboratory, Oak Ridge, Tennessee, United States of America; Email: birriah@ornl.gov.

**Rajni Chahal -** Oak Ridge National Laboratory, Chemical Science Division, Oak Ridge, Tennessee, United States of America; orcid.org/0000-0003-2190-0473; Email: chahalr@ornl.gov.

**Ibrahim Isah -** University of Massachusetts Lowell, Department of Chemical Engineering, 1 University Ave, SOU-202E, Lowell, Massachusetts, United States of America; Email: ibrahim_isah@student.uml.edu.

**Muhammad Nouman Anwar -** University of Massachusetts Lowell, Lowell, Massachusetts, United States of America; Email: muhammadnouman_anwar@student.uml.edu.

**Tyler Starkus -** Argonne National Laboratory, Lemont, Illinois, United States of America; Email: tstarkus@anl.gov.

**Prasanna Balaprakash -** Oak Ridge National Laboratory, Computing and Computational Sciences Directorate, Oak Ridge, Tennessee, United States of America; Email: pbalapra@ornl.gov.

**Acknowledgements**

J.B., R.C., I.I, M.N.A., and S. L. acknowledge funding from the National Science Foundation, award number 2138456.

**Competing interests**

The authors declare no competing interests.

**Additional information**

Supplementary information. The online version contains supplementary material available at [link provided at the moment of publication].